\documentclass[final]{cvpr}

\usepackage{times}
\usepackage{epsfig}
\usepackage{amsmath,amssymb}
\usepackage[utf8]{inputenc}
\usepackage{diagbox}
\usepackage{graphicx}
\usepackage{amsmath}
\usepackage{amssymb}
\usepackage{booktabs}
\usepackage{subfigure}
\usepackage{float}

\usepackage[dvipsnames]{xcolor}
\definecolor{g}{HTML}{12B208} 
\definecolor{p}{HTML}{db7093}
\usepackage{multirow}
\usepackage[T1]{fontenc}

\usepackage[pagebackref=true,breaklinks=true,colorlinks,bookmarks=false]{hyperref}

\def\cvprPaperID{46} 
\def\confYear{CVPR 2021}

\begin{document}

\title{Distill on the Go: Online knowledge distillation in self-supervised learning}

\author{Prashant Bhat, Elahe Arani, Bahram Zonooz \\
Advanced Research Lab, NavInfo Europe, Eindhoven, The Netherlands \\
{\tt\small \{prashant.bhat, elahe.arani\}@navinfo.eu}, {\tt\small bahram.zonooz@gmail.com} \\
}

\maketitle

\begin{abstract}
    Self-supervised learning solves pretext prediction tasks that do not require annotations to learn feature representations. For vision tasks, pretext tasks such as predicting rotation, solving jigsaw are solely created from the input data. Yet, predicting this known information helps in learning representations useful for downstream tasks. However, recent works have shown that wider and deeper models benefit more from self-supervised learning than smaller models. To address the issue of self-supervised pre-training of smaller models, we propose Distill-on-the-Go (DoGo), a self-supervised learning paradigm using single-stage online knowledge distillation to improve the representation quality of the smaller models. We employ deep mutual learning strategy in which two models collaboratively learn from each other to improve one another. Specifically, each model is trained using self-supervised learning along with distillation that aligns each model’s softmax probabilities of similarity scores with that of the peer model. We conduct extensive experiments on multiple benchmark datasets, learning objectives, and architectures to demonstrate the potential of our proposed method. Our results show significant performance gain in the presence of noisy and limited labels, and in generalization to out-of-distribution data.
\end{abstract}

\section{Introduction}
Self-supervised learning (SSL) has recently begun to rival the performance of supervised learning on computer vision tasks \cite{byol, chen2020simple, swav}. SSL learns meaningful representations from data without requiring manually annotated labels. To learn task-agnostic visual representations, SSL solves pretext prediction tasks such as predicting relative position \cite{doersch2015unsupervised} and/or rotation \cite{gidaris2018unsupervised}, solve jigsaw \cite{kim2018learning} and image in-painting \cite{pathak2016context}. Predicting known information helps in learning representations generalizable for downstream tasks such as segmentation and object detection \cite{lee2020predicting}. 


Recent works have empirically shown that deeper and wider models benefit more from task agnostic use of unlabeled data than their smaller counterparts i.e smaller models when trained using self-supervised learning fail to close in the gap with respect to supervised training \cite{fang2021seed, chen2020big}. For instance level discrimination tasks such as contrastive learning, smaller models with fewer parameters are hard to optimize given large amounts of data. The limitation in the representation quality lies in the difficulty of optimization rather than the model size \cite{jimmyBa2014}. 

\begin{figure}
\centering
\includegraphics[width=0.99\linewidth]{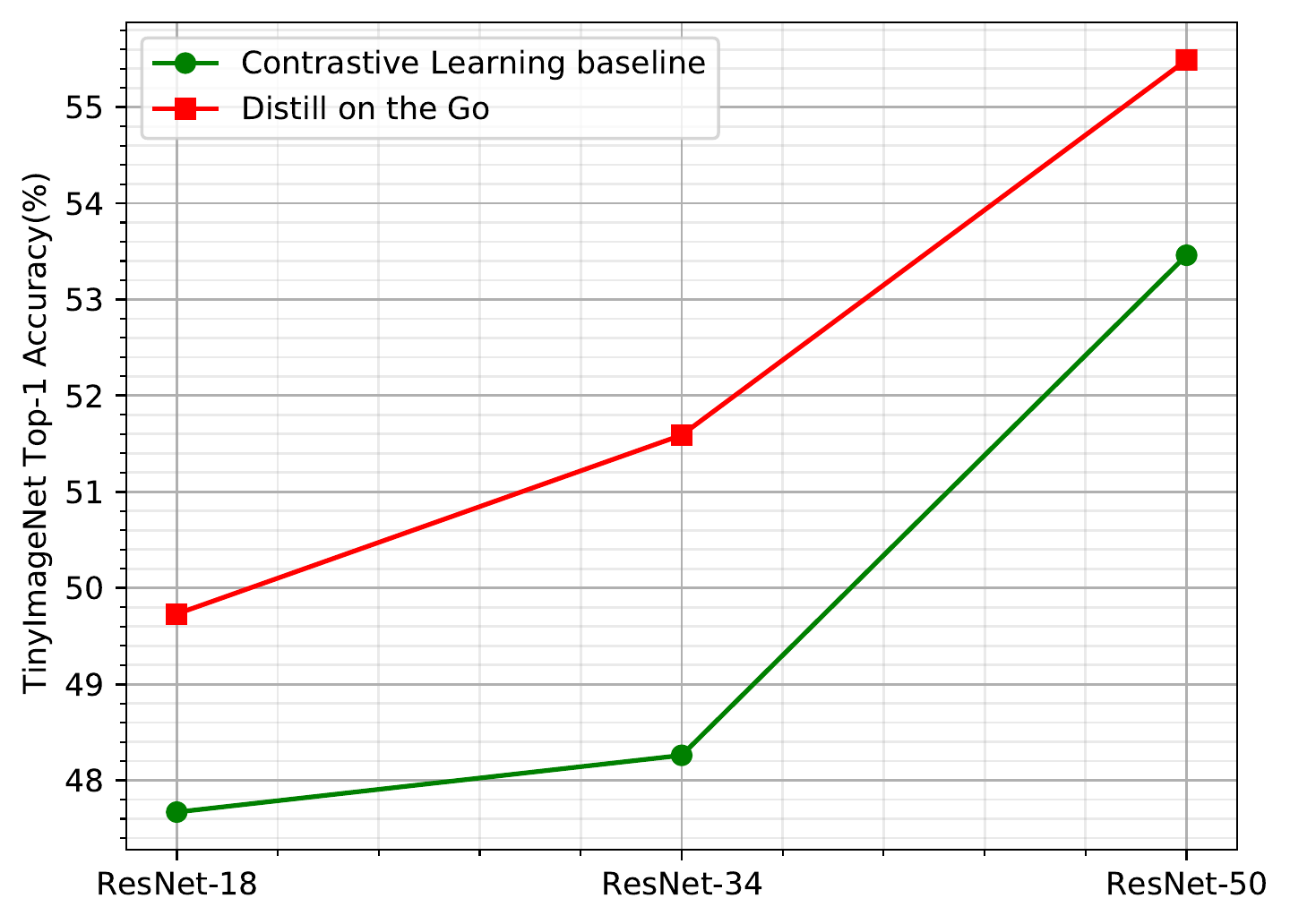} 
\caption{Tiny-ImageNet evaluation: comparison of state-of-the-art self-supervised contrastive learning against our proposed method Distill-on-the-go using linear evaluation. Self-supervised models are trained using SimCLR while Distill-on-the-go models are trained together with ResNet-50.}
\label{fig:g_matrices}
\end{figure}

To address this issue and improve the representation quality of smaller models, we leverage knowledge distillation (KD) \cite{hinton2015distilling}. KD involves training a student model under the supervision of a larger pre-trained teacher model in an interactive manner, similar to how humans learn. Many of the traditional KD methods formulate the student to mimic the softened softmax output of the teacher. Thus, these methods are not directly applicable to self-supervised knowledge distillation in the absence of the labels. Also, offline KD methods 
require longer training process and significantly large memory and computational resources to pre-train large teacher models \cite{onthefly2018}.  

Online knowledge distillation offers a more attractive alternative owing to its one stage training and bidirectional knowledge distillation \cite{zhang2017deep, Qiushan2020, onthefly2018, sarfraz2021knowledge}. These approaches treat all (typically two) participating models equally, enabling them to learn from each other. In this paper, we propose a fully self-supervised online knowledge distillation framework, \textit{Distill-on-the-go (DoGo)}. In contrast to offline knowledge distillation, our proposed method starts with multiple untrained models which simultaneously learn by solving a pretext task. Specifically, we train two models collaboratively by applying different augmentation for each model using our method. Each model generates two projections corresponding to two randomly augmented views. We then align temperature scaled similarity scores across these projections for KD in which the relative differences in similarity between the reference sample and other samples in a batch can provide additional useful information. The additional supervision signal can assist the optimization of the smaller model. Furthermore, the collaboration between multiple models enables them to explore the feature space more extensively and aids in converging to a more robust (flatter) minima which leads to a better generalization to unseen data \cite{zhang2017deep}.


With extensive experiments on different datasets, we empirically demonstrate that learning along with peers is better than learning alone in a conventional self-supervised learning setting. For example, on Tiny-ImageNet dataset \cite{Le2015TinyIV}, our method improves top-1 accuracy(\%) of ResNet-50 by $2.7\%$ and ResNet-18 by $4.5\%$ compared to contrastive learning baselines (Figure \ref{fig:g_matrices}). 
Our main contributions are as follows: 
\begin{itemize}
    \item We propose a fully self-supervised online KD method DoGo to improve the representation quality of the smaller models. The method is completely self-supervised, i.e. knowledge is distilled during the pre-training stage in the absence of labels. 
    \item With DoGo, we demonstrate a significant improvement in performance for smaller models over contrastive learning baselines.
    \item We demonstrate the efficacy of DoGo on multiple SSL algorithms, thereby showing that our method is independent of the underlying SSL learning objective.
    \item We further show the effectiveness of DoGo in solving more common challenges in the real-world problems including learning with noisy and limited labels, and generalization to out-of-distribution data.
\end{itemize}


\section{Related Work}
\textbf{Self-supervised learning: }SSL can be broadly categorized into generative and contrastive methods \cite{liu2020selfsupervised}. Generative self-supervised models try to learn meaningful visual representations by re-constructing either a part of an input or a whole of it. For example, Auto-Regressive models use observations from previous time step to predict the joint distribution factorised as a product of conditionals. PixelRNN \cite{prnn} and PixelCNN \cite{pixelCNN} adapt this idea to model image pixel by pixel.

Contrastive learning, on the other hand, learns to compare through Noise Contrast Estimation \cite{nce}. Context-instance contrast methods model belonging relationship between local feature of a sample and its global context. Predict Relative Position (PRP) and Maximize Mutual Information (MI) fall under this category. In PRP, self-supervised methods focus on learning visual representations by predicting relative position of local components. PRP pretext tasks include predict relative position \cite{doersch2015unsupervised} and/or rotation \cite{gidaris2018unsupervised}, solve jigsaw \cite{kim2018learning} etc. In MI, Deep InfoMax \cite{hjelm2018learning} employs Mutual Information Neural Estimation \cite{belghazi2018mutual} to learn unsupervised representations by simultaneously maximizing and estimating mutual information between learned high level representations and the input data.

Although MI based contrastive self-supervised methods achieved quite a bit of success, some recent methods outperform them through context-context contrastive learning \cite{liu2020selfsupervised}. These methods discard mutual information, instead learn the relationship between the global representations of different samples akin to metric learning. InstDisc \cite{wu2018unsupervised} proposed instance discrimination as a pretext task. CMC \cite{cmc} employed multi-view contrastive learning framework with multiple different views of an image as positive samples and take views of other images as the negatives. MoCo \cite{MoCo} further developed the idea of instance discrimination by leveraging momentum contrast. MoCo addressed two critical issues in dealing with the negative sampling: (i) Momentum contrast prevents the fluctuation of loss convergence in the earlier stages, (ii) MoCo maintains a queue to store negative samples from previous batches thereby significantly improving the negative sample efficiency. However, MoCo adopts too simple augmentations, thus making positive pair far too easy to identify. Hard positive sample strategy plays a key role in instance discrimination as outlined in SimCLR \cite{chen2020simple}. Therefore, SimCLR relinquishes momentum contrast overall but retains the siamese structure and introduces augmentations of 10 forms with an end-to-end training framework. SimCLRv2 \cite{chen2020big} outlined that bigger models benefit more from a task agnostic use of unlabeled data for visual representation learning. Owing to larger modeling capacity, bigger self-supervised models are far more label efficient and perform better than smaller models on downstream tasks.

\begin{figure*}[hbt]
\centering
\includegraphics[trim=0 7 0 7 cm, width=1\linewidth]{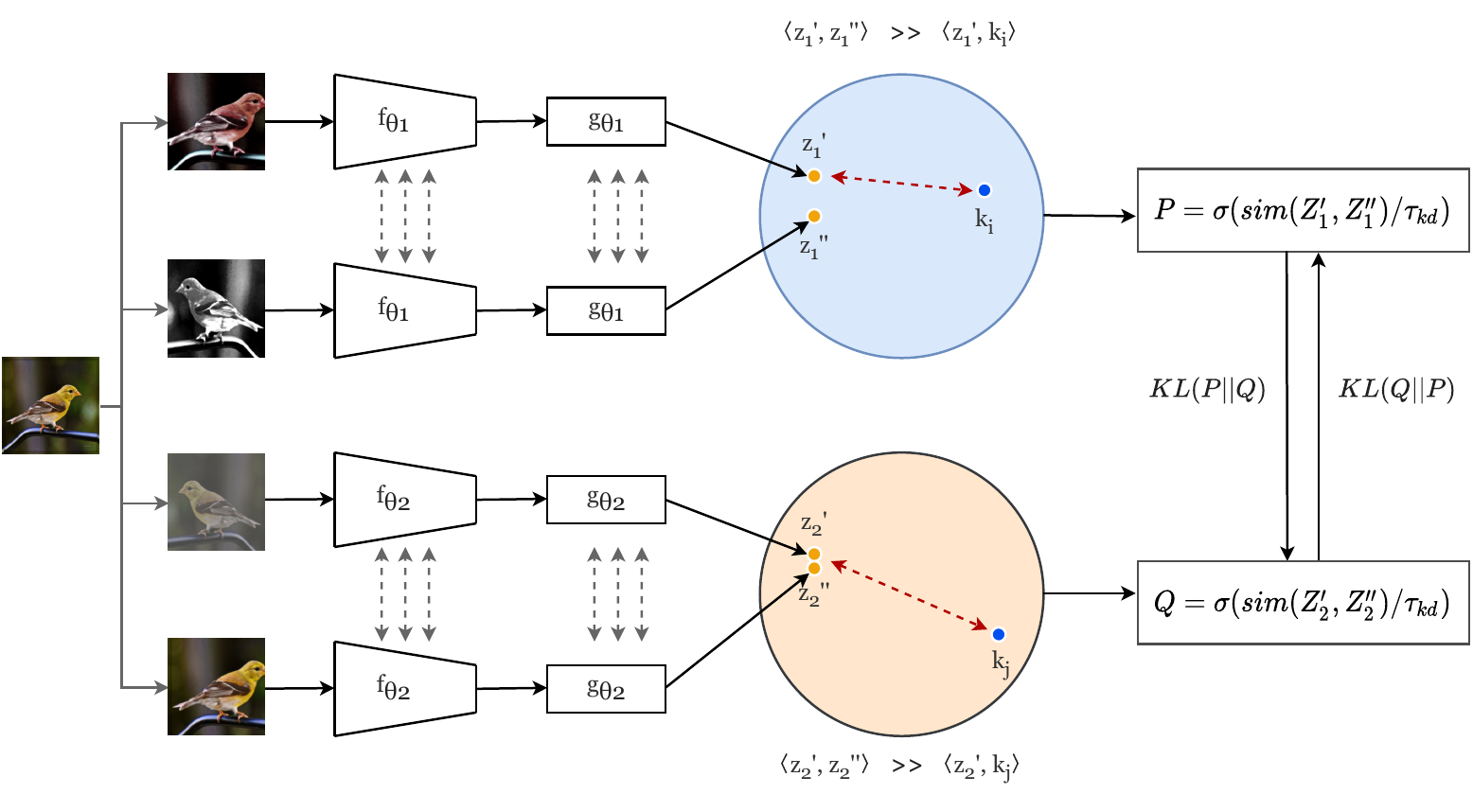}
\caption{Distill-on-the-go: The proposed method consists of a stochastic augmentation module, followed by shared encoder and a two-layer MLP projection head. The participating models primarily learn visual representations by contrastive loss. $\langle ., . \rangle$ represents cosine similarity between two embeddings. The extra learning experience is provided by way of aligning softmax probabilities  of  temperature-scaled  similarity  scores across augmentations of two peer models. }
\label{fig:short}
\end{figure*}

\textbf{Knowledge distillation: }KD is an effective technique for improving the performance of compact models either using the supervision of larger pre-trained model or with a cohort of smaller models trained collaboratively.
In a typical ``teacher-student" knowledge distillation, smaller student model is trained under the supervision of a pre-trained larger teacher model, producing a higher representation quality compared to training student model from scratch. In the original formulation, Hinton \etal \cite{hinton2015distilling} proposed a representation distillation by way of mimicking softened softmax output of the teacher. Better generalization can be achieved by emulating the latent feature space in addition to mimicking the output of the teacher \cite{sarfraz2020knowledge, sp, fsp, fitnet, rkd}. 


To circumvent the associated computational costs of pre-training a teacher, deep mutual learning (DML) \cite{zhang2017deep} proposed online knowledge distillation using Kullback–Leibler (KL) divergence. Alongside a primary supervised cross-entropy loss, DML involves training each participating model using a distillation loss that aligns the class posterior probabilities of the current model with that of the other models in the cohort. Inspired by the recent progress made by online knowledge distillation methods, we propose a fully self-supervised  online knowledge distillation method to improve the representation quality of the smaller models. To the best of our knowledge, we are the first to propose an online knowledge distillation method in the absence of labels.

\section{Proposed Method}
We begin with the core idea of unsupervised contrastive representation learning. We wish to learn visual representations by contrasting semantically similar (positive) and dissimilar (negative) pairs of data samples such that similar pairs have the maximum agreement via a contrastive loss in the latent space. Contrastive loss encourages the representations of similar pairs to be close, while those of dissimilar pairs to be more orthogonal in the latent space \cite{chuang2020debiased}. Empirically, similar pairs are obtained by randomly augmenting the same sample using rotation, gaussian noise and color jittering \cite{chen2020simple, wu2018unsupervised, MoCo}. 

Inspired by the recent advancements in contrastive representation learning, DoGo comprises of the following major components: It consists of a stochastic augmentation module resulting in two highly correlated views $I'$ and $I''$ of the same input sample $I$. 
The correlated views are then fed into $f_{\theta}(.)$, typically an encoder network such as ResNet-50 \cite{resnets} and subsequently to $g_{\theta}(.)$, a two-layer perceptron with ReLU non-linearity. To learn the visual representations, the network $g_{\theta}(f_{\theta}(.))$ should learn to maximize the similarity between the positive embedding pair $\langle z', z'' \rangle$ while simultaneously pushing away the negative embedding pairs $\langle z', k_{i} \rangle$, where $i = (1,...., K)$ are the embeddings of augmented views of other samples in a batch and $K$ is the number of negative samples. Contrastive representation learning can thus be cast as an instance level discrimination task. Instance level discrimination objective is typically formulated using a softmax criterion. However, the cost of computing non-parametric softmax is prohibitively large especially when the number of instances are very large \cite{wu2018unsupervised}. Popular techniques to reduce computation include hierarchical softmax \cite{hierarchical}, noise contrast estimation \cite{nce} and negative sampling \cite{negative_sampling}. Following \cite{chen2020simple, chen2020big},  We use noise contrast estimation for a positive embedding pair $\langle z'_i, z''_i \rangle$ where $i\in\{1,2\}$ indicates the two models as follows:
\begin{equation}
\label{eq:contrastive_loss}
L_{cl,i} =  - log\frac { { e }^{ sim(z'_i,z''_i)/\tau_{c}  } }{ { e }^{ sim(z'_i,z''_i)/\tau_{c}  }+\sum _{ j=1 }^{ K }{ { e }^{ sim(z'_i,k_j)/\tau_{c}  } }  } 
\end{equation}
\noindent $L_{cl}$ is a normalized temperature scaled cross entropy loss \cite{chen2020simple}. Wang \etal \cite{Wang_2017} provided an in-depth understanding of necessity of normalization when using dot product of feature vectors in a cross entropy loss. Therefore, we use cosine similarity (l2 normalized dot product) in the computation of the contrastive loss $L_{cl}$. 

\begin{table*}[t!h]
\centering
\begin{tabular}{c|c|c|c}\hline
 \toprule
 
Baseline & ResNet-50 (53.46) & ResNet-34 (48.26) & ResNet-18 (47.58) \\
\midrule

ResNet-50 (53.46) & 55.49 \, \textbackslash \, 54.93 & 55.24 \, \textbackslash \, 51.59 & 54.93 \, \textbackslash \, 49.73 \\
& \textcolor{g}{+3.8\% \hspace{0.33cm} +2.7\%} & \textcolor{g}{+3.3\% \hspace{0.33cm} +6.9\%} & \textcolor{g}{+2.7\% \hspace{0.33cm} +4.5\%} \\
\midrule

ResNet-34 (48.26) & 51.59 \, \textbackslash \, 55.24 & 50.86 \, \textbackslash \, 51.32 & 50.26 \, \textbackslash \, 49.10 \\
& \textcolor{g}{+6.9\% \hspace{0.33cm} +3.3\% } & \textcolor{g}{+5.4\% \hspace{0.33cm} +6.3\%} & \textcolor{g}{+4.1\% \hspace{0.33cm} +3.1\%} \\

\midrule

ResNet-18 (47.67) & 49.73 \, \textbackslash \, 54.93 & 49.10 \, \textbackslash \, 50.26 & 48.77 \, \textbackslash \, 48.74 \\
& \textcolor{g}{+4.5\% \hspace{0.33cm} +2.7 } & \textcolor{g}{+3.1\% \hspace{0.33cm} +4.1\%} & \textcolor{g}{+2.3\% \hspace{0.33cm} +2.2\%} \\
\bottomrule \bottomrule
\\

Baseline & WRN-28-2 (37.10) & WRN-16-2 (34.28) & WRN-10-2 (30.75) \\
\midrule     

WRN-28-2 (37.10) & 38.16 \, \textbackslash \, 37.73 & 37.79 \, \textbackslash \, 36.30 & 36.49 \, \textbackslash \, 33.55 \\
& \textcolor{g}{+2.9\% \hspace{0.33cm} +1.7\%} & \textcolor{g}{+1.9\% \hspace{0.33cm} +5.9\%} & \textcolor{p}{-1.7\%} \hspace{0.33cm}  \textcolor{g}{+9.1\%} \\
\midrule  

WRN-16-2 (34.28) & 36.30 \, \textbackslash \, 37.79 & 35.92 \, \textbackslash \, 35.68 & 34.69 \, \textbackslash \, 33.04 \\
& \textcolor{g}{+5.9\% \hspace{0.33cm} +1.9\% } & \textcolor{g}{+4.8\% \hspace{0.33cm} +4.1\%} & \textcolor{g}{+1.2\% \hspace{0.33cm} +7.4\%} \\
\midrule

WRN-10-2 (30.75) & 33.55 \, \textbackslash \, 36.49 & 33.04 \, \textbackslash \, 34.69 & 32.53 \, \textbackslash \, 32.60 \\
& \textcolor{g}{+9.1\%} \hspace{0.33cm} \textcolor{p}{-1.7\%} & \textcolor{g}{+7.4\% \hspace{0.33cm} +1.2\%} & \textcolor{g}{+5.8\% \hspace{0.33cm} +6.0\%} \\
\bottomrule
\end{tabular}
\caption{Tiny-ImageNet top-1 accuracy(\%) under linear evaluation for various ResNet and Wide-ResNet models. Baseline measures the linear evaluation accuracy when the models are pre-trained using contrastive learning alone. Each box contains DoGo results: left value corresponds to model in the corresponding row, right value corresponds to model in the corresponding column. Percentage change in accuracy between baseline and our method are highlighted in green.}
\label{tab:main}
\end{table*}

Smaller models find it hard to optimize and find right set of parameters in instance level discrimination tasks, attributing to difficulty of optimization rather than the model size. We believe that the additional supervision in KD regarding the relative differences in similarity between the reference sample and other sample pairs and the collaboration between multiple models can assist the optimization of the smaller model. Therefore to improve generalizability of smaller model $g_{\theta 1}(f_{\theta 1}(.))$, we propose to utilize another peer model $g_{\theta 2}(f_{\theta 2}(.))$. Given a new sample, each participating peer model generates embeddings $ z', z''$ of two different augmented views. Let $Z', Z'' \in R^{N \times m}$ be a batch of $ z', z''$ where $N$ is batch size and $m$ is the length of the projection vector. Let $P = \sigma (sim(Z'_{1}, Z''_{1}) / \tau)$ and $Q = \sigma (sim(Z'_{2}, Z''_{2}) / \tau)$ be softmax probabilities of temperature-scaled similarity scores across augmentations of two peer models. We employ KL divergence to distill the knowledge across peers by aligning the distributions $P$ and $Q$. The distillation losses are defined as follows:
\begin{equation} 
\label{eq:kd}
\begin{split}
    L_{kd, 1} &= D_{KL}(Q || P) \\
&= \sigma \left(\frac{sim(Z'_{2}, Z''_{2})}{\tau_{kd}}\right)  log  \frac{ \sigma (sim(Z'_{2}, Z''_{2}) / \tau_{kd})}{ \sigma (sim(Z'_{1}, Z''_{1}) / \tau_{kd})}
\end{split}
\end{equation}
\begin{equation} 
\label{eq:kd}
\begin{split}
    L_{kd, 2} &= D_{KL}(P || Q) \\
&= \sigma \left(\frac{sim(Z'_{1}, Z''_{1})}{\tau_{kd}}\right)  log  \frac{ \sigma (sim(Z'_{1}, Z''_{1}) / \tau_{kd})}{ \sigma (sim(Z'_{2}, Z''_{2}) / \tau_{kd})}
\end{split}
\end{equation}

The final learning objective for the two participating models can be written as:
\begin{equation} 
\label{eq:final}
L_{\theta_1} = L_{cl,1} + \lambda L_{kd,1}
\end{equation}
\begin{equation} 
\label{eq:final}
L_{\theta_2} = L_{cl,2} + \lambda L_{kd,2}
\end{equation}
where $\lambda$ is a regularization parameter for adjusting the magnitude of the knowledge distillation loss. Our method can also be extended to more than two peers by simply computing distillation loss with all the peers. Another strategy is to create an ensemble distribution from all participating peers and use it as a single teacher for creating learning experience. We reserve training with more than two peers as a possible future research direction.

\section{Experimental Results}

\subsection{Implementation details}
We use SimCLR \cite{chen2020simple} as our baseline network. It consists of a feature extractor and an additional projector. The projector is a two layer MLP with ReLU non-linearity generating embeddings of size 128. Since Tiny-ImageNet images are much smaller than ImageNet, we replace the first $7\times7$ Conv of stride 2 with a $3\times3$ Conv of stride 1 in ResNet backbones. We also remove max pooling layer from the first Conv block. Our stochastic augmentation module employs random resized crop, random horizontal flip followed by random color distortions. Since images are small in size, we leave out the Gaussian blur so as not to risk over-augmentation. We use Adam optimizer with a fixed learning rate of $3e^{-4}$ and weight decay of $1e^{-6}$ with a batch size of 256 across all pre-training experiments. To make a fair comparison, we use $\lambda=100$ keeping other hyperparameters intact in knowledge distillation experiments.

\textbf{Datasets:} We use Tiny-ImageNet \cite{Le2015TinyIV} as our primary dataset for self-supervised pre-training. It has 200 classes and 500 samples per class each $64\times64$ in size. The test set contains 10,000 images. Tiny-ImageNet pre-trained models are evaluated on six different out-of-distribution datasets including CIFAR-10 \cite{cifar}, CIFAR-100 \cite{cifar}, STL-10 \cite{stl-10} and three DomainNet \cite{DomainNet} datasets. DomainNet is a domain adaptation dataset with six domains and $0.6$ million images distributed across $345$ categories. We have used ClipArt, Sketch and QuickDraw domain datasets in this work. 

\subsection{Evaluation metrics}

\textbf{Linear Evaluation:} 
To evaluate the learned representations, we follow the widely used linear evaluation metric \cite{chen2020simple, cpc} as our primary evaluation protocol. A linear layer is trained on the labeled training set on top of frozen encoder $f_{\theta}(.)$. Top-1 test accuracy(\%) is used as a quality measure for learned representations. To reduce hyperparameter tuning per experiment, we standardize the linear evaluation experiments as follows: We use Adam optimizer with a fixed learning rate of $3e^{-4}$ and weight decay of $1e^{-6}$ with a batch size of 64. All linear evaluations are run for 100 epochs except out-of-distribution experiments, which are evaluated for 50 epochs. 

\begin{table}[t!h]
\centering
\begin{tabular}{ccc}
\toprule
Label(\%) & CL Baseline & DoGo\\
 \midrule
1 & 15.00 & \textbf{15.46}  \\
\midrule
5 & 29.49 & \textbf{29.64} \\
\midrule
10 & 34.23 & \textbf{36.07} \\
\midrule
20 & 40.23 & \textbf{42.43} \\
\midrule
50 & 49.81 & \textbf{50.46} \\ 
\midrule
100 & 54.76 & \textbf{55.57} \\
\bottomrule
\end{tabular}
\caption{Tiny-ImageNet top-1 accuracy(\%) for ResNet-18 with varied available labels during fine-tuning. Labels are sampled in a class balanced manner. CL baseline is trained using contrastive loss while DoGo is jointly trained with ResNet-50. DoGo consistently outperforms the contrastive learning baseline.}
\label{tab:finetune}
\end{table}

\textbf{Nearest Neighbor Evaluation:} 
We also evaluate learned representations using nearest neighbour classification with an efficient similarity search library Faiss GPU \cite{faiss}. We extract the features from frozen encoder $f_{\theta}(.)$ and index features along with their labels. We report the average of nearest neighbour classification with $k=1, 2, 4, 8$ neighbors. The results can be found in the supplementary material.

\subsection{Evaluation on in-distribution data} \label{linear_eval}
Table \ref{tab:main} compares the top-1 accuracy(\%) of our proposed method on Tiny-ImageNet dataset obtained by ResNet and Wide-ResNet architectures. Following observations  can be made from the table: (i) All participating models under ResNet family consistently see an improvement over the contrastive learning baseline. (ii) Although ResNet-50 is a much larger network compared to  ResNet-18, ResNet-50 still sees a significant improvement. It is therefore clear that all participating models can benefit from joint training unless there is a drastic difference in modeling capacity. (iii) Wide-ResNets also display a similar trend with an exception of WRN-28-2 owing to a large difference in model capacity between WRN-10-2 and WRN-28-2. However, WRN-10-2 achieves a significant improvement of over $9\%$. (iv) Smaller models benefit more from joint training than their larger counterparts. Given the difficulty of training smaller models under contrastive learning regime, joint training under our proposed method can significantly improve their performance. (v) Smaller models gain a consistent improvement with the increase in modeling capacity of their counterparts. For example, the performance gain in ResNet-18 monotonically increases as we increase the capacity of the peer model from ResNet-18 to ResNet-34 and ResNet-50. A similar trend can be seen for WRN-10-2, WRN-16-2 and ResNet-34. (vi) Contrary to conventional wisdom, even joint training of two models with exact same architecture yields a discernible improvement over the baseline.

\begin{table*}[hbt!]
\centering
\bigskip
\begin{tabular}{ll|c|c|c|c|c|c}
\toprule
 && CIFAR-10 & CIFAR-100 & STL-10 & Clip Art & Quick Draw & Sketch\\
 \midrule
\multirow{2}{*}{ResNet-50}  & CL Baseline & 78.59 & 53.54 & 76.78 & 37.82 & 47.31 & 31.04 \\
& DoGo & \textbf{80.26} & \textbf{56.36} & \textbf{77.93} & \textbf{42.20} & \textbf{51.88} & \textbf{33.62} \\
\midrule

\multirow{2}{*}{ResNet-18} & CL Baseline & 72.50 & 45.98 & 72.90 & 29.10 & 39.97 & 23.68 \\
& DoGo & \textbf{75.19} &\textbf{ 48.52} & \textbf{74.73} & \textbf{33.51} & \textbf{43.43} & \textbf{26.25} \\
\bottomrule
\end{tabular}
\caption{Linear evaluation on out-of-distribution datasets. CL baseline is trained using contrastive loss while DoGo is a joint training of ResNet-50 and ResNet-18.}
\label{tab:ood}
\end{table*}

\begin{figure}[tb!h]
\centering
\includegraphics[width=0.99\linewidth]{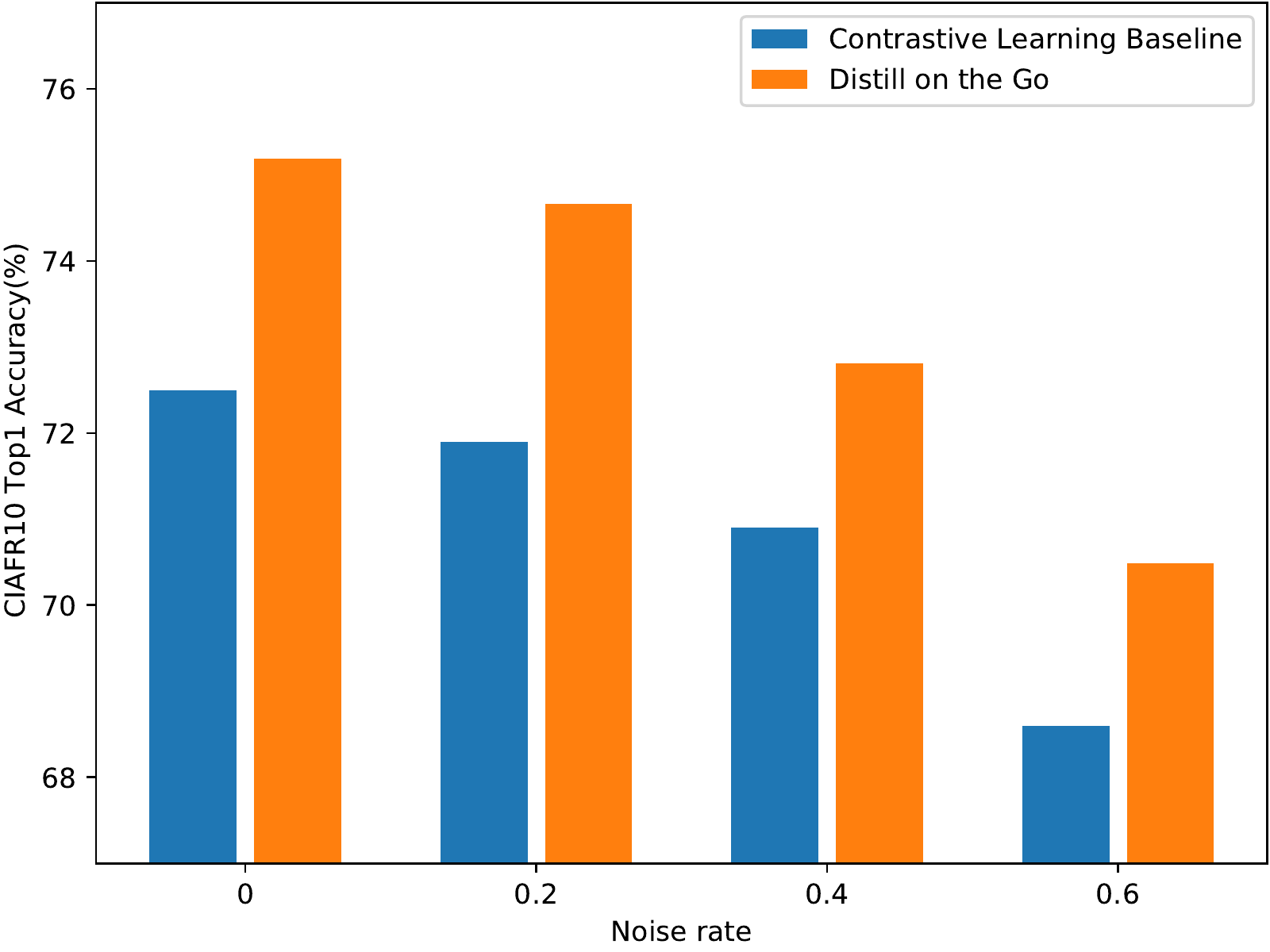}
\caption{CIFAR-10 top-1 accuracy(\%) for ResNet-18 under linear evaluation with simulated noisy labels. CL baseline is trained using contrastive learning while DoGo is jointly trained with ResNet-50. DoGo significantly outperforms the CL baseline across all noise rates.}
\label{fig:noisy}
\end{figure}

\subsection{Evaluation under varying label quantities}
We sample 1\%, 5\%, 10\%, 20\%, 50\% and 100\% of the Tiny-ImageNet in a class balanced manner and simply fine-tune the whole encoder $f_{\theta}(.)$ with a linear layer on top on the labeled data. We use 15\% of the training data as a validation set and top-1 accuracy(\%) are reported on full test set. Table \ref{tab:finetune} shows the comparison of fine-tuning pre-trained models on fewer labels against contrastive learning baseline. It can be clearly seen that our method outperforms the contrastive learning baseline across all limited label scenarios. The difference is more lucid when the available labels are 10\% or higher. The difference is higher in the lower mid range (10\%, 20\%) and saturates thereafter. It is therefore clear that online knowledge distillation brings discernible benefits when pre-trained models are fine-tuned on the downstream task.

\subsection{Generalization on out-of-distribution data}
Self-supervised contrastive learning learns the visual representations in a task agnostic manner. It is therefore essential that these learned representations generalize well across downstream tasks. To evaluate the models pre-trained on Tiny-ImageNet, we consider six different out-of-distribution datasets. Table \ref{tab:ood} compares the top-1 test accuracy(\%) under linear evaluation protocol of pre-trained models with and without online knowledge distillation. ResNet-50 and ResNet-18 generalize well across all datasets when trained together using our method compared to the baseline. Results show that DoGo generalizes well across datasets with drastically different distribution than Tiny-ImageNet indicating DOGo learns better feature representations. 

\begin{figure*}[htb]
\centering
\subfigure[ResNet-18]{\includegraphics[width=0.495\linewidth]{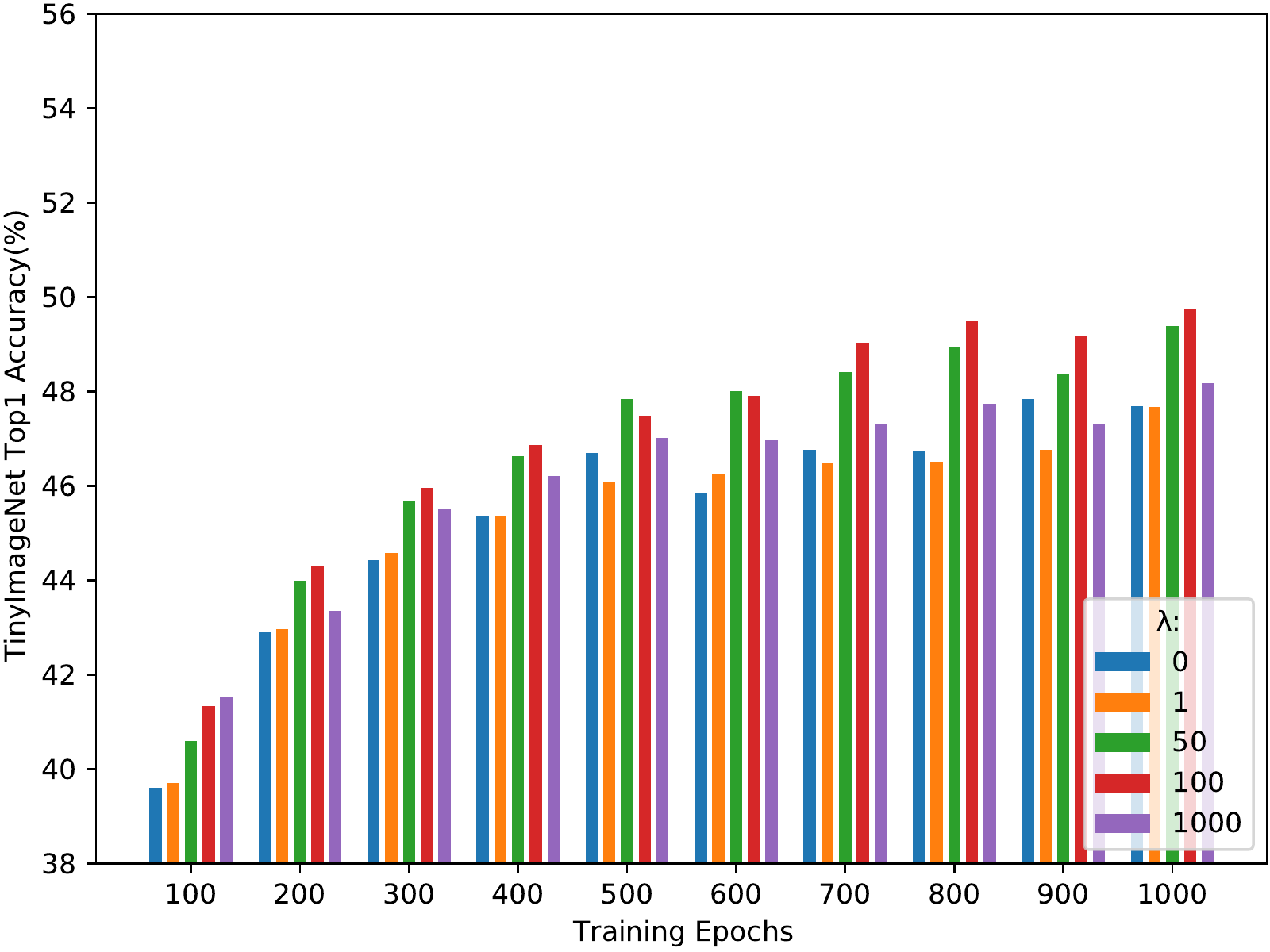}}
\subfigure[ResNet-50]{\includegraphics[width=0.495\linewidth]{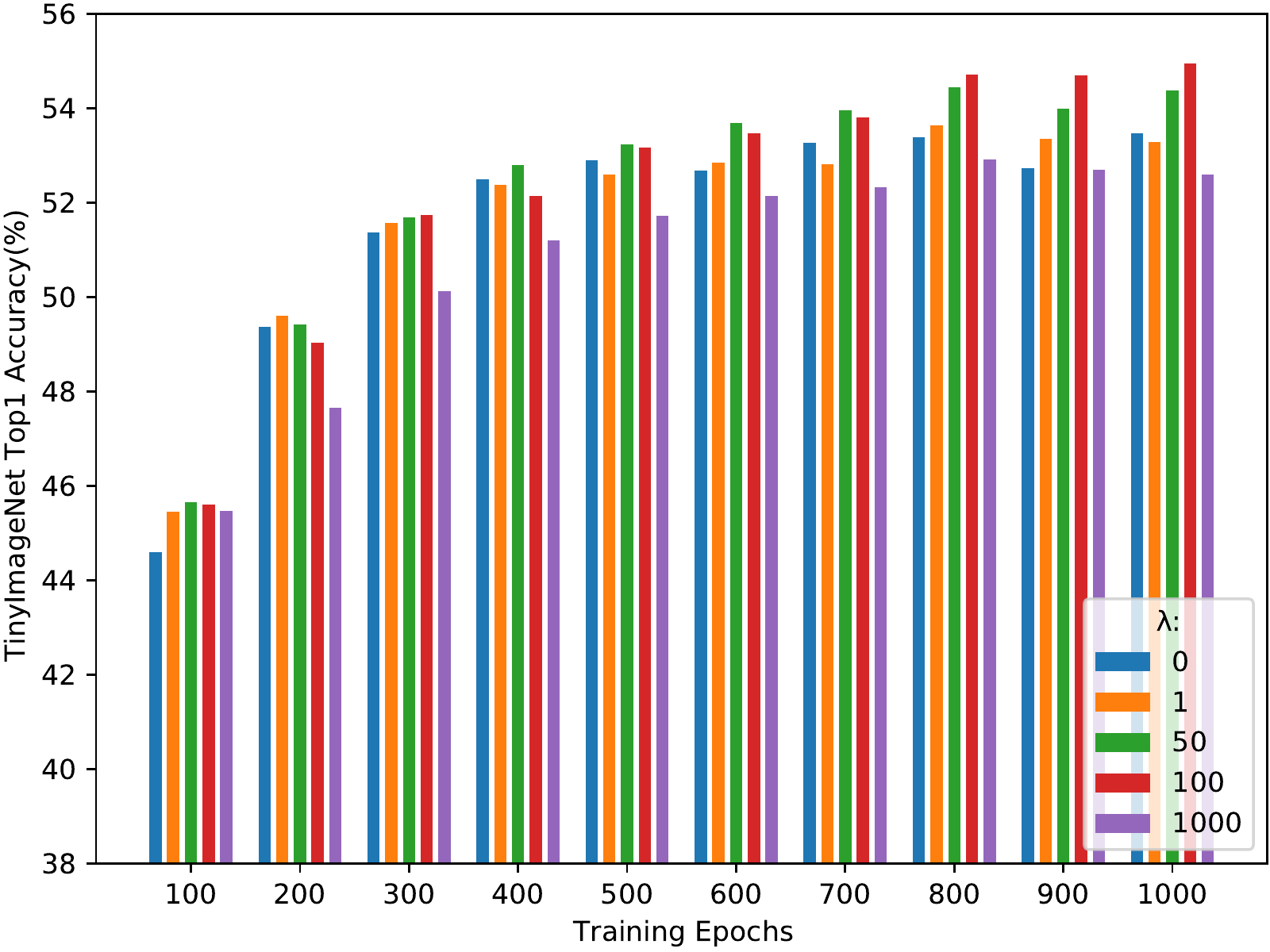}}
\caption{Hyperparameter tuning of regularization parameter $\lambda$}
\label{fig:ablation_lambda}
\end{figure*}

\subsection{Evaluation under noisy labels}
The success of supervised learning often hinges on the availability of huge amounts of high quality annotations. Quite often, large datasets contain noisy labels due to difficulty of manual annotation. It is pertinent to have robust training procedures to offset the impact of noisy labels as studies have shown that deep neural networks can memorize noisy labels \cite{arpit2017closer}. Self-supervised learning decouples representation learning and classifier, thus, is more robust to noisy labels than supervised learning \cite{hendrycks2019using, zhang2020decoupling}. We hypothesize that learned representations in DoGo are more robust to noisy labels. To test our hypothesis, we simulate label corruption on CIFAR-10. With a given probability (noise rate), we corrupt every true label by randomly sampling from a uniform distribution over number of classes. Figure \ref{fig:noisy} presents the results of linear evaluation under noisy labels. Our method consistently outperforms the contrastive learning baseline across different noise rates.

\begin{table}[bt!h]
\centering
\begin{tabular}{c|c|c}
\toprule
\multirow{2}{*}{Baseline} & Contrastive loss $L_{cl}$ & SimSiam loss $L_{ss}$\\ 
& (47.67) & (46.80) \\
 \midrule
 $L_{cl}$ & \textbf{48.77 \, \textbackslash \, 48.74} &  \textbf{48.61 \, \textbackslash \, 49.03}   \\
 (47.67) & \textcolor{g}{+2.3\% \hspace{0.33cm} +2.2\%} & \textcolor{g}{+2.0\% \hspace{0.33cm} +4.8\%} \\
 \midrule
 $L_{ss}$ & \textbf{49.03 \, \textbackslash \, 48.61}  &  \textbf{48.95 \, \textbackslash \, 48.38}  \\
 (46.80) & \textcolor{g}{+4.8\% \hspace{0.33cm} +2.0\%} & \textcolor{g}{+4.6\% \hspace{0.33cm} +3.4\%} \\
\bottomrule
\end{tabular}
\caption{Tiny-ImageNet top-1 accuracy(\%) for ResNet-18 under linear evaluation for different learning objectives. DoGo is jointly trained with a ResNet-18 peer model. Each box contains DoGo results: left value corresponds to model in the corresponding row, right value corresponds to model in the corresponding column. Percentage change in accuracy between baseline and our method are highlighted in green. DoGo surpasses the baselines on all combinations of learning objectives.}
\label{tab:simsiam}
\end{table}

\subsection{Distillation across learning objectives}
The experimental results shown thus far have had contrastive learning as their primary learning objective. In this section, we explore whether it is possible to distill knowledge across peers with different learning objective. In contrast to contrastive learning, SimSiam \cite{simsiam} is one such learning objective that can learn meaningful visual representations without negative sample pairs and large batches. SimSiam can best be described as BYOL \cite{byol} without momentum encoder, SimCLR \cite{chen2020simple} without negative pairs and Swav \cite{swav} without online clustering. Essentially, it covers all leading self-supervised learning  methods and makes an ideal candidate for our research. In addition to encoder $f_{\theta}(.)$ and projection MLP $g_{\theta}(.)$, SimSiam has a prediction MLP $h_{\theta}(.)$ that transforms the output of one view and matches it to the other view. Denoting  two output vectors $p' = h_{\theta }(g_{\theta }(f_{\theta }(I')))$ and  $z'' = g_{\theta }(f_{\theta }(I''))$, the learning objective is to minimize the negative cosine similarity,
\begin{equation} 
\label{eq:simsiam}
L_{ss} = D(p', z'') = - \frac{p'}{\left | \left | p' \right | \right |_{2}} \cdot \frac{z''}{\left | \left | z'' \right | \right |_{2}}
\end{equation}

To evaluate our method across learning objectives, we train one model using contrastive loss (Eq. \ref{eq:contrastive_loss}) and the other by SimSiam loss (Eq. \ref{eq:simsiam}). Since SimSiam model has both projections ($\in R^{b \times 2048}$) and predictions ($\in R^{b \times 2048}$), we use predictions for creating learning experience during joint training. Table \ref{tab:simsiam} shows the generalizability of our method beyond contrastive learning. Our method surpasses the baseline across all combinations of learning objectives. Results reinforce our earlier proposition that DoGo is independent of underlying SSL objective.

\begin{figure}[tb!h]
\centering
\includegraphics[width=0.75\linewidth]{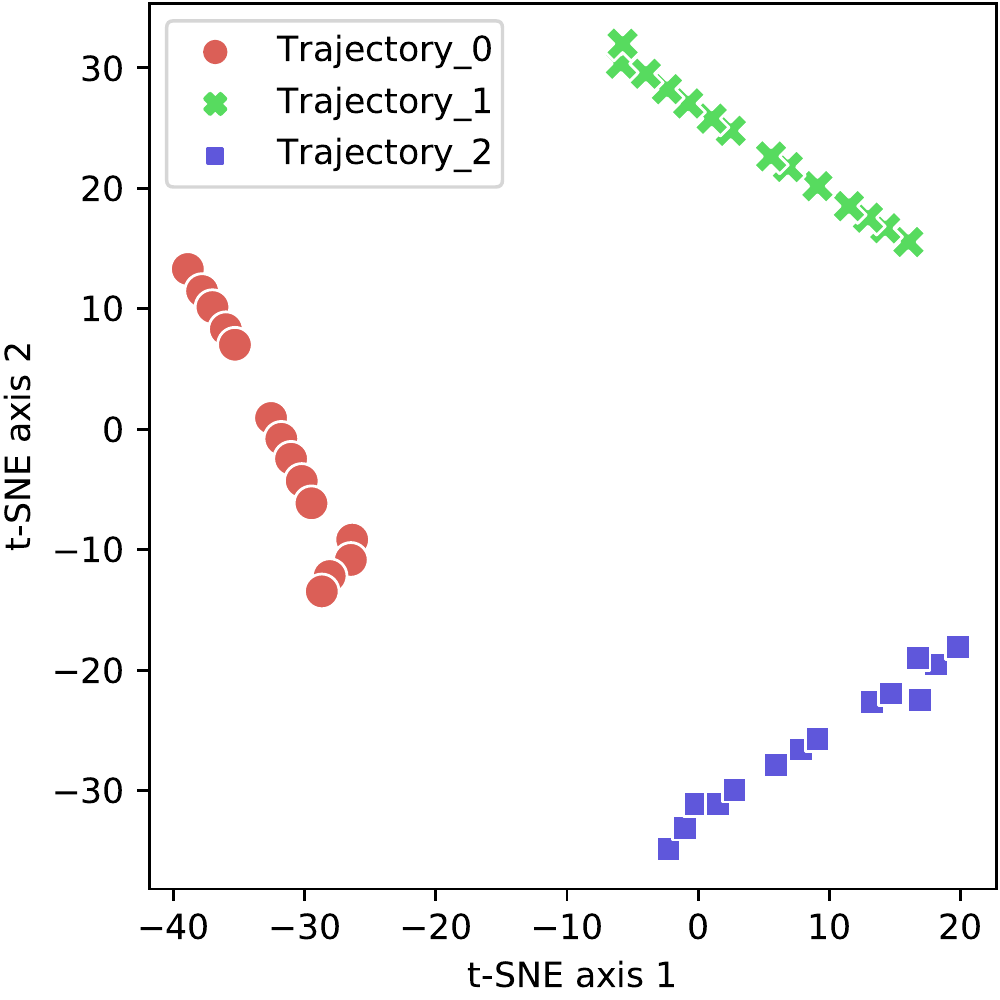}
\caption{t-SNE plot of projections from checkpoints corresponding to 3 different randomly initialized trajectories.}
\label{fig:tsne}
\end{figure}

\begin{figure*}[hbt!]
  \centering
  \subfigure[Temperature in $L_{cl}$]{\includegraphics[width=0.49\linewidth]{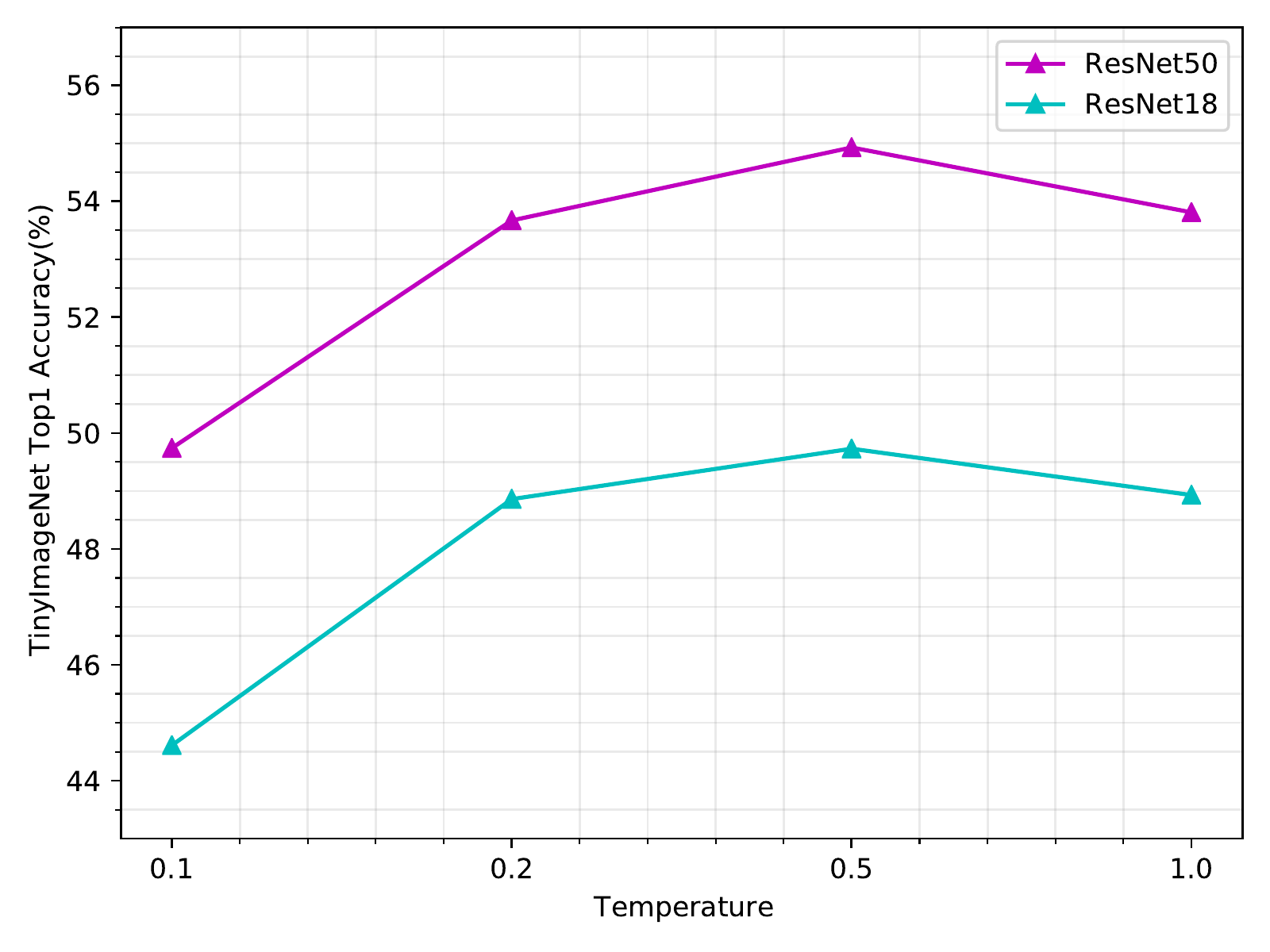}}
  \subfigure[Temperature in $L_{kd}$]{\includegraphics[width=0.49\linewidth]{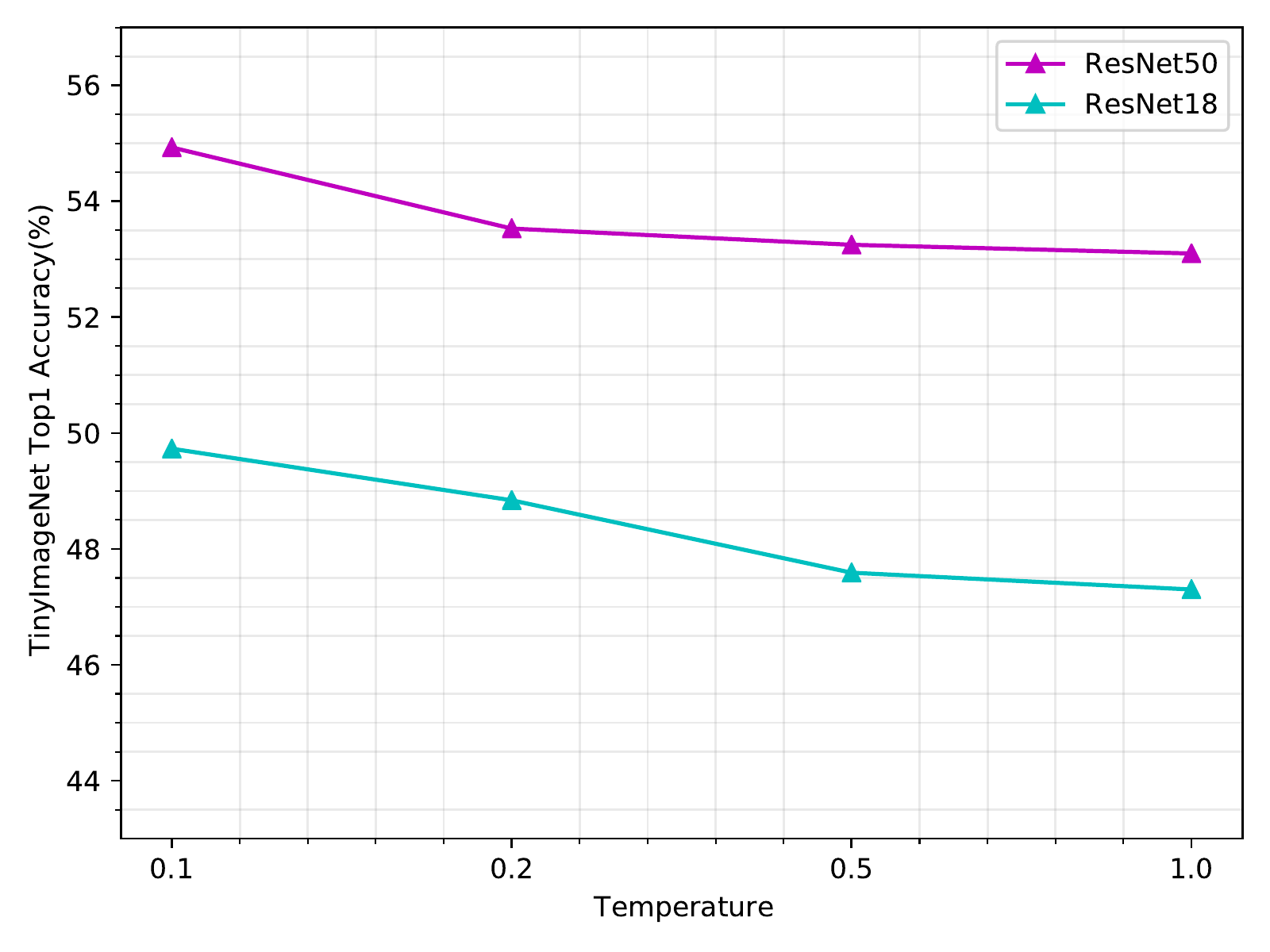}}
  \caption{Hyperparameter tuning of scalar temperature parameters (a) $\tau_{c}$ in contrastive loss and (b) $\tau_{kd}$ in knowledge distillation loss }
\label{fig:temp}
\end{figure*}

\subsection{Visualizing function space similarity}
In this section, we attempt to provide additional intuition about why our method works. Motivated by \cite{fort2020deep}, we plot t-SNE \cite{tsne} of projections of different checkpoints along the individual training trajectories from multiple initializations on Tiny-ImageNet trained with contrastive learning. Trajectory\_i in Figure \ref{fig:tsne} shows the variation in the function space during the training of model i. We can see that a single model only explores a limited function space whereas models initialized differently explore considerably different function spaces even though they provide similar performance. Therefore, we believe that the collaborative learning between multiple models in DoGo enables the models to collectively explore broader region of the function space resulting in more generalizable feature representations. DoGo enables the exchange of information across different family of functions resulting in improved downstream performance.



\section{Effect of hyperparameters}
We now explore the influence of $\lambda$ and scalar temperature parameter $\tau$ on knowledge distillation.

\textbf{Regularization parameter \pmb{$\lambda$}:} Figure \ref{fig:ablation_lambda} summarizes the effect of $\lambda$ parameter when ResNet-18 and ResNet-50 are trained together using our method. We draw several conclusions from the graph: (i) We see a clear performance improvement as the training progresses i.e. online knowledge distillation reaps more benefits from longer training. This is in line with  similar empirical results on self-supervised learning \cite{chen2020simple}. (ii) Smaller models benefit early in the training while larger counterparts see a performance improvement in the later stages of the training. The smaller model ResNet-18, sees an improvement early in the training while ResNet-50 is benefited only in the later stages. We attribute this behaviour to large modeling capacity difference between these two models. (iii) For both ResNet-18 and ResNet-50, $\lambda=1$ yields either comparable or slightly worse performance. This is due to the difference in magnitudes of contrastive loss and distillation loss functions. To compensate, we increase the weight $\lambda$ for the distillation loss. We find that $\lambda=100$ yields optimum performance for both participating models. Increasing $\lambda$ further is detrimental for the downstream performance.

\textbf{Temperature parameters \pmb{$\tau_{c}$} and \pmb{$\tau_{kd}$}: }Several prior works have shown that an appropriately adjusted temperature parameter benefits contrastive cross entropy loss \cite{chen2020simple, kalantidis2020hard, wu2018unsupervised}. We can see from Eq. \ref{eq:contrastive_loss} that the range of similarity can be scaled using an appropriate temperature value. Since the gradients of Eq. \ref{eq:contrastive_loss} will be scaled by $1/\tau_{c}$, tuning this parameter could effectively tune the hardness of the negative samples and speed of learning \cite{kalantidis2020hard}. In our implementation, we use two decoupled scalar temperature parameters $\tau_{c}$ and $\tau_{kd}$. Figure \ref{fig:temp} shows the linear evaluation of ResNet-18 and ResNet-50 trained together using online knowledge distillation. We find that when training to convergence, the optimal $\tau_{c}$ is $0.5$ and $\tau_{kd}$ is $0.1$. The optimal value of $\tau_{c}=0.5$ is in accordance with the empirical results in contrastive learning. We also find that decoupling of $\tau_{c}$ and $\tau_{kd}$ yields a much better result in our experiments.

\section{Conclusion and future work}
We proposed Distill-on-the-Go, a self-supervised learning paradigm using single stage online knowledge distillation to improve the representation quality of the smaller models. Through extensive experiments across multiple network architectures, SSL algorithms and datasets, we demonstrated the efficacy of DoGo over self-supervised learning baselines resulting in compact yet accurate models. Performance of compact models can be boosted even further by training them with much larger models. We further demonstrated the additional benefits of using online distillation in SSL in enabling the models to learn efficiently under varying degrees of label noise and different levels of annotations data availability. DoGo also enables the models to explore the function space more extensively and learn better generalizable features.
Joint training with more than two peers, online knowledge distillation on bigger datasets such as ImageNet, downstream performance on other tasks such as object detection and segmentation are some of the useful future research directions for this work.

{\small
\bibliographystyle{ieee_fullname}
\bibliography{sample}
}

\newpage

\appendix
\section{Nearest Neighbor Evaluation}

Table \ref{tab:nearest_neighbor} presents the nearest neighbor evaluation of ResNet and Wider ResNet models on Tiny-ImageNet dataset. The observations made under section 4.3 still hold. All participating models except for WRN10-2 show an improvement over baseline. Smaller models benefit more from joint training than their larger counterparts. Also, their performance improves with the increase in modeling capacity of their counterparts.

\begin{table*}[hbt!]
\centering
\begin{tabular}{c|c|c|c}\hline
 \toprule
 
Baseline & ResNet-50 (33.07) & ResNet34 (32.07) & ResNet-18 (29.90) \\

\midrule
ResNet-50 (33.07) & 34.4 \, \textbackslash \, 34.76 & 34.70 \, \textbackslash \, 34.06 & 34.05 \, \textbackslash \, 31.49 \\
& \textcolor{g}{+4.0\% \hspace{0.33cm} +5.1\%} & \textcolor{g}{+4.9\% \hspace{0.33cm} +6.2\%} & \textcolor{g}{+3.0\% \hspace{0.33cm} +5.3\%} \\

\midrule

ResNet34 (32.07) & 34.06 \, \textbackslash \, 34.70 & 33.12 \, \textbackslash \, 33.96 & 32.57 \, \textbackslash \, 31.51 \\
& \textcolor{g}{+6.2\% \hspace{0.33cm} +4.9\% } & \textcolor{g}{+3.3\% \hspace{0.33cm} +5.9\%} & \textcolor{g}{+1.6\% \hspace{0.33cm} +5.4\%} \\
\midrule

ResNet-18 (29.90) & 31.49 \, \textbackslash \, 34.05 & 31.51 \, \textbackslash \, 32.57 & 30.82 \, \textbackslash \, 30.86 \\
& \textcolor{g}{+5.3\% \hspace{0.33cm} +3.0\% } & \textcolor{g}{+5.4\% \hspace{0.33cm} +1.6\%} & \textcolor{g}{+3.0\% \hspace{0.33cm} +3.2\%} \\
\bottomrule \bottomrule
\\

Baseline & WRN28-2 (22.03) & WRN16-2 (19.30) & WRN10-2 (17.20) \\
\midrule  

WRN28-2 (22.03) & 22.77 \, \textbackslash \, 22.44 & 22.14 \, \textbackslash \, 21.15 & 21.82 \, \textbackslash \, 19.76 \\
& \textcolor{g}{+3.4\% \hspace{0.33cm} +1.9\%} & \textcolor{g}{+0.4\% \hspace{0.33cm} +9.6\%} & \textcolor{p}{-0.9\%} \hspace{0.33cm}  \textcolor{g}{+14.9\%} \\
\midrule  

WRN16-2 (19.30) & 21.15 \, \textbackslash \, 22.14 & 20.97 \, \textbackslash \, 20.42 & 20.20 \, \textbackslash \, 19.59 \\
& \textcolor{g}{+9.6\% \hspace{0.33cm} +0.4\% } & \textcolor{g}{+8.7\% \hspace{0.33cm} +5.8\%} & \textcolor{g}{+4.6\% \hspace{0.33cm} +13.9\%} \\
\midrule

WRN10-2 (17.20) & 19.76 \, \textbackslash \, 21.82 & 19.59 \, \textbackslash \, 20.2 & 19.17 \, \textbackslash \, 19.52 \\
& \textcolor{g}{+14.9\%} \hspace{0.33cm} \textcolor{p}{-0.9\%} & \textcolor{g}{+13.9\% \hspace{0.33cm} +4.6\%} & \textcolor{g}{+11.5\% \hspace{0.33cm} +13.5\%} \\
\bottomrule

\end{tabular}
\caption{Tiny-ImageNet top-1 accuracy(\%) under nearest neighbor evaluation for various ResNet and Wider ResNet models.  Baseline measures the linear evaluation accuracy when the models are pre-trained using contrastive learning alone.  Each box contains DoGo results:  left value corresponds to model in the corresponding row, right value corresponds to model in the corresponding column.  Percentage change inaccuracy between baseline and our method are highlighted in green.}
\label{tab:nearest_neighbor}
\end{table*}

\end{document}